\documentclass[twocolumn,english]{article}
\usepackage{lmodern}
\usepackage[T1]{fontenc}
\usepackage[latin9]{inputenc}
\usepackage{geometry}
\usepackage{color}
\usepackage{babel}
\usepackage{verbatim}
\usepackage{amsmath}
\usepackage{amssymb}
\usepackage{esint}
\usepackage[authoryear]{natbib}
\usepackage[unicode=true,pdfusetitle,
 bookmarks=true,bookmarksnumbered=false,bookmarksopen=false,
 breaklinks=false,pdfborder={0 0 0},backref=false,colorlinks=true]
 {hyperref}
\hypersetup{
 citecolor=blue}
\usepackage{breakurl}

\makeatletter

\providecommand{\tabularnewline}{\\}

\usepackage{subfigure} 
\usepackage{graphicx} 

\DeclareMathOperator*{\argmin}{arg\,min}

\usepackage{tikz}
\usetikzlibrary{bayesnet}

\usepackage{algorithm}
\usepackage{algorithmic}

\usepackage{natbib}

\makeatother

\begin{document}

\title{\textbf{\huge{}Simple approximate MAP Inference for Dirichlet processes}}

\author{Yordan P. Raykov, Alexis Boukouvalas, Max A. Little}

\date{October 27, 2014}

\maketitle
\global\long\def\N{\mathcal{N}}
\global\long\def\x{{\bf x}}
\global\long\def\vmu{\mu}
\global\long\def\Nk{N_{k}}
\global\long\def\samplemean{\bar{\x}}
\global\long\def\z{z}
\global\long\def\Nkn{N_{k,-i}}
\global\long\def\conc{\alpha}
\global\long\def\Z{\mathcal{Z}}
\global\long\def\X{\mathbf{X}}
\global\long\def\thp{\theta_{0}}
\global\long\def\R{\mathbf{R}}
\global\long\def\St{St}

\global\long\def\vsp{\vspace{0.1cm}}
\global\long\def\N{\mathcal{N}}
\global\long\def\x{{\bf x}}
\global\long\def\vmu{\mu}
\global\long\def\Nk{N_{k}}
\global\long\def\th{\theta}
\global\long\def\m{{\bf m}}
\global\long\def\b{{\bf b}}

\begin{abstract}
The Dirichlet process mixture (DPM) is a ubiquitous, flexible Bayesian
nonparametric statistical model. However, full probabilistic inference
in this model is analytically intractable, so that computationally
intensive techniques such as Gibb's sampling are required. As a result,
DPM-based methods, which have considerable potential, are restricted
to applications in which computational resources and time for inference
is plentiful. For example, they would not be practical for digital
signal processing on embedded hardware, where computational resources
are at a serious premium. Here, we develop simplified yet statistically
rigorous approximate maximum a-posteriori (MAP) inference algorithms
for DPMs. This algorithm is as simple as $K$-means clustering, performs
in experiments as well as Gibb's sampling, while requiring only a
fraction of the computational effort. Unlike related small variance
asymptotics, our algorithm is non-degenerate and so inherits the ``rich
get richer'' property of the Dirichlet process. It also retains a
non-degenerate closed-form likelihood which enables standard tools
such as cross-validation to be used. This is a well-posed approximation
to the MAP solution of the probabilistic DPM model.
\end{abstract}

\section{Introduction}

Bayesian nonparametric (BNP) models have been successfully applied
on a wide range of domains but despite significant improvements in
computational hardware, statistical inference in most BNP models remains
infeasible in the context of large datasets. The flexibility gained
by such models is paid for with severe decreases in computational
efficiency, and this makes these models somewhat impractical. Therefore,
there is an emerging need for approaches that simultaneously minimize
both empirical risk and computational complexity \citep{bousquet2008tradeoffs}.
Towards that end we present a simple, statistically rigorous and computationally
efficient approach for the estimation of BNP models based on maximum
a-posteriori (MAP) inference and concentrate on inference in \emph{Dirichlet
process mixtures }(DPMs).

DPMs are mixture models which use the \emph{Dirichlet Process} (DP)
\citep{ferguson1973} as a prior over the parameters of the distribution
of some random variable. The random variable has a distribution with
a potentially infinite number of mixture components. The DP is an
adaptation of the discrete Dirichlet distribution to the infinite,
uncountable sample space. A draw from a DP is itself a density function.
A DP is the Bayesian conjugate density to the empirical probability
density function, much as the discrete Dirichlet distribution is conjugate
to the categorical distribution. Hence, DPs have value in Bayesian
probabilistic models because they are priors over completely general
density functions. This is one sense in which DPMs are nonparametric.

An additional, interesting property of DP-distributed density functions
is that they are discrete in the following sense: they are formed
of an infinite, but countable mixture of Dirac delta functions. Since
the Dirac has zero measure, the support of the density function is
also countable. This discreteness means that draws from such densities
have a non-zero probability of being repeats of previous draws. Furthermore,
the more often a sample is repeated, the higher the probability of
that sample being drawn again -- an effect known as the ``rich get
richer''\emph{ }property (known as \emph{preferential attachment
}in the network science literature \citep{Barabasi1999}). This repetition,
coupled with preferential attachment, leads to another valuable property
of DPs: samples from DP-distributed densities have a strong \emph{clustering
}property whereby $N$ draws can be partitioned into $K$ representative
draws, where $K<N$ and $K$ is not fixed \textit{a-priori}.

Practical methods to deal with drawing densities from DPs, and samples
from these densities, revolve around three equivalent constructions:
the (generalized) \emph{Pólya urn }scheme \citep{blackwell1973ferguson}
allowing draws of samples directly from the DP, the \emph{stick-breaking
}method \citep{ishwaran2001gibbs} which creates explicit representations
of DP-distributed densities, and the \emph{Chinese restaurant process}
(CRP) which defines exchangeable conditional distributions over partitions
of the draws from the DP, defined by the $K$ representatives. All
three constructions lead to practical stochastic sampling inference
schemes for DPMs. Sampling inference is most often conducted using
Markov Chain Monte Carlo (MCMC) methods, e.g. Gibbs \citep{Geman1984}
and slice \citep{neal2003} samplers. Here we discuss only the Gibbs
sampler as it is the ``starting point'' for the development of the
novel algorithm of this paper.

Approximate inference algorithms that scale better with data size
include \emph{variational Bayes }(VB) schemes for DPMs \citep{blei2004variational}.
However, the factorization assumption required leads to biased estimates
of the posterior of interest. Further, VB algorithms can become trapped
in local minima and often underestimate the variance of the quantities
of interest \citep[page 462]{Bishop2006prml}. VB for DPMs also truncate
the infinite number of components in the DPM which causes additional
approximation error. More importantly, even with this truncation,
closed form maximization steps for the different DPM quantities are
rarely obtained so that iterative optimization steps are required.
Although DPM VB often converges more quickly than MCMC, it usually
requires high computational effort for each step. Finally, obtaining
the optimization schedule for the variational distribution with respect
to the auxiliary variables involved is often a challenging task.

\emph{Small variance asymptotics} create degenerate, point mass Dirac
distributions in the probabilistic model to devise simplified inference
algorithms. The \emph{DP-means} algorithm \citep{ICML2012Kulis} is
derived from a Gaussian DPM Gibbs sampler by shrinking the component
covariances to zero (further discussed in Section \ref{sub:Hard-clustering-using}).
The approach was later extended to exponential family distributions
by \citet{Kulis2013}. Later \citet{Broderick2013} uses DP-means
as a tool for finding the MAP solution of the degenerate complete
likelihood for a DPM, and applies the same principle to Bayesian nonparametric\emph{
}latent feature models. However, in using a degenerate likelihood,
some of the defining properties of the DPM, for example the prior
over the partition (see Section \ref{sub:Chinese-Restaurant-Process}),
are lost. In this work, we present an algorithm for finding the solution
of the MAP problem posed in \citet{Broderick2013} without resorting
to a degenerate likelihood. This enables the algorithm to be more
faithful to inference in the corresponding probabilistic model, and
also allows the use of standard rigorous tools such as out-of-sample
prediction for cross-validation.

We concentrate on inference in DP mixtures and show how the CRP may
be exploited to produce simplified MAP inference algorithms for DPMs.
Similar to DP-means it provides only point estimates of the joint
posterior. However, while DP-means follows the close relationship
between $K$-means and the finite Gaussian mixture model (GMM) to
derive a ``nonparametric $K$-means'', we exploit the concept of
\emph{iterated conditional modes} (ICM) \citep{kittler1984contextual}.

After reviewing the CRP (Section \ref{sub:Chinese-Restaurant-Process})
and DPM (Section \ref{sub:The-Dirichlet-Process}), we discuss the
DP-means algorithm highlighting some of its deficiencies in Section
\ref{sub:Hard-clustering-using} and we show how these can be overcome
using non-degenerate MAP inference in Section \ref{sec:MAP-DP-description}.
We compare the different approaches on synthetic and real datasets
in Section \ref{sec:Experiments-on-Synthetic} and we conclude with
a discussion of future directions Section \ref{sec:Discussion-and-Extensions}.

\section{Background}

\subsection{Chinese restaurant process (CRP)\label{sub:Chinese-Restaurant-Process}}

The CRP is a discrete time stochastic process over the sample space
of partitions, or equivalently can be thought as a probability distribution
over cluster indicator variables. It is strictly defined by an integer
$N$ (number of items) and a positive, real \emph{concentration }parameter
$\alpha$. A draw from a CRP has probability:
\begin{equation}
p\left(z_{1},\ldots,z_{N}\right)=\frac{\Gamma\left(\alpha\right)}{\Gamma\left(N+\alpha\right)}\alpha^{K}\prod_{k=1}^{K}\Gamma\left(N_{k}\right)\label{eq:joint over z}
\end{equation}

with indicators $z_{1},\ldots,z_{N}\in\left\{ 1,\ldots,K\right\} $,
where $K$ is the unknown number of items and $N_{k}=\left|\left\{ i:z_{i}=k\right\} \right|$
is the number of indicators taking value $k$ with $\sum_{k=1}^{K}N_{k}=N$.
For any finite $N$ we will have $K\le N$ and usually $K$ will be
much smaller than $N$, so the CRP returns a partitioning of $N$
elements into some smaller number of groups $K$. The probability
over indicators is constructed in a sequential manner using the following
conditional probability: 
\begin{equation}
p\left(z_{n+1}=k\left|z_{1},\ldots,z_{n}\right.\right)=\begin{cases}
\frac{N_{k}}{\alpha+N} & \text{if }k=1,\ldots,K\\
\frac{\alpha}{\alpha+N} & \text{otherwise}
\end{cases}\label{eq:conditional for CRP}
\end{equation}
By increasing the value of $n$ from $1$ to $N$ and using the corresponding
conditional probabilities, we obtain the joint distribution over indicators
from (\ref{eq:joint over z}), $p\left(z_{1},\ldots,z_{N}\right)=p\left(z_{N}\left|z_{1},\ldots,z_{N-1}\right.\right)p\left(z_{N-1}\left|z_{1},\ldots,z_{N-2}\right.\right)\cdots p\left(z_{2}\left|z_{1}\right.\right)$.
The stochastic process is often explained using the metaphor of customers
sitting at tables at a Chinese restaurant, where the probability of
customer $n+1$ sitting on a previously occupied table or a new table
is given by (\ref{eq:conditional for CRP}).

\subsection{The Dirichlet process Gaussian mixture model and Gibbs sampler\label{sub:The-Dirichlet-Process}}

DPMs are popular nonparametric Bayesian models, related to the finite
mixture model, but making additional assumptions that allow for greater
flexibility. For illustration, consider the case where the mixture
components are Gaussian with joint mean and precision parameters $\left(\boldsymbol{\mu},\mathbf{R}\right)$.
We will denote using $\mathbf{X}$ the full data matrix formed of
the observed data points $\boldsymbol{\x}_{i}$ which are $D$-dimensional
vectors $\boldsymbol{\x}_{i}=\left(x_{i,1},\ldots,x_{i,d},\ldots,x_{i,D}\right)$.
The \emph{Dirichlet process Gaussian mixture model} (DPGMM) with collapsed
mixture weights can be written: 
\begin{eqnarray}
\zeta_{N} & \sim & \textrm{CRP}\left(\alpha,N\right)\nonumber \\
\left(\boldsymbol{\mathbf{\mu}}_{k},\mathbf{\R}_{k}\right) & \sim & G_{0}\label{eq:generative model DPGMM-1}\\
\x_{i} & \sim & \mathcal{N}\left(\boldsymbol{\mathbf{\mu}}_{z_{i}},\mathbf{\R}_{z_{i}}^{-1}\right)\nonumber 
\end{eqnarray}

for $k=1,\ldots,K$. Here, $\zeta_{N}$ denotes a partition drawn
from a CRP which implicitly constrains the indicators for $N$ data
points. Variables $\boldsymbol{\mu}_{k}$ and $\mathbf{\R}_{k}$ are,
respectively, component means and precision matrices that are jointly
\emph{normal-Wishart} (NW) distributed. Here, $i=1,2,3\ldots,N$ indexes
observations $\boldsymbol{\x}_{i}$ and $G_{0}$ is a prior density
function over these parameters. Given the clustering property of the
CRP, there will be $K$ draws $\left(\boldsymbol{\mu}_{k},\mathbf{R}_{k}\right)$
from $G_{0}$ pointed\emph{ }to \emph{by cluster indicators $\left\{ i:z_{i}=k\right\} $,
}so that $\left(\boldsymbol{\mu}_{k},\mathbf{R}_{k}\right)$ become
the cluster parameters for cluster $k$. Each observation is then
Gaussian but sharing parameters with all the other observations in
the same cluster i.e. $\mathcal{N}\left(\boldsymbol{\x}_{i}\left|\boldsymbol{\mu}_{k},\mathbf{R}_{k}^{-1}\right.\right)$
for $z_{i}=k$.

One simple and popular sampling algorithm for parameter inference
in the DPM is \emph{CRP-based Gibbs sampling}, first applied by \citet{West1994}
and later discussed in \citet{neal2000markov}. This MCMC algorithm
alternates until convergence between the two stages of sampling the
component indicators while holding the cluster parameters fixed, and
sampling new cluster parameters while holding the cluster indicators
fixed. The first stage samples the cluster indicators $z_{i}$ from
the following conditional distribution:

\begin{equation}
p\left(z_{i}=k\left|\boldsymbol{\x}_{i},\boldsymbol{\mu}_{k},\mathbf{R}_{k}\right.\right)\propto N_{k,-i}\mathcal{N}\left(\boldsymbol{\x}_{i}\left|\boldsymbol{\mu}_{k},\mathbf{R}_{k}^{-1}\right.\right)
\end{equation}

where $N_{k,-i}$ denotes the number of times an observation, apart
from the observation $\boldsymbol{\x}_{i}$, has been assigned to
cluster $k$. However, as this is a DPM there is always a finite probability
of creating a new cluster, whereby we sample new cluster parameters
$\left(\boldsymbol{\mu}_{K+1},\mathbf{R}_{K+1}\right)$ and add a
new value $K+1$ to the possible values for each indicator. The probability
of creating a new cluster is:

\begin{equation}
p\left(z_{i}=K+1\left|\boldsymbol{\x}_{i},G_{0}\right.\right)\propto\alpha\int\mathcal{N}\left(\boldsymbol{\x}_{i}\left|\boldsymbol{\mu},\mathbf{R}^{-1}\right.\right)dG_{0}(\boldsymbol{\mu},\mathbf{R})
\end{equation}

The parameters $\left(\boldsymbol{\mu}_{K+1},\mathbf{R}_{K+1}\right)$
are sampled from the posterior distribution defined by the prior $G_{0}$
and the single observation likelihood $\mathcal{N}\left(\boldsymbol{\x}_{i}\left|\boldsymbol{\mu},\mathbf{R^{-1}}\right.\right)$.
Due to conjugacy, this posterior is NW with the prior parameters we
have set for $G_{0}$. In the second stage, we sample the new cluster
parameters $\left(\boldsymbol{\mu}_{k},\mathbf{R}_{k}\right)$ for
each cluster $k=1,\ldots,K$ from the posterior distribution over
$\left(\boldsymbol{\mu}_{k},\mathbf{R}_{k}\right)$ conditional on
the updated $z_{i}$, the prior $G_{0}$, and the joint likelihood
for all observations assigned to cluster $k$, i.e. every $\boldsymbol{\x}_{i}$
where $z_{i}=k$. This likelihood is a product of $N_{k}$ Gaussian
distributions, and by conjugacy, the posterior distribution will be
another NW.

\subsection{Hard clustering via small variance asymptotics\label{sub:Hard-clustering-using}}

Based on the CRP Gibbs sampler described above and with some simplifying
assumptions, \citet{ICML2012Kulis} describe a \emph{hard clustering
}algorithm that is closely related to $K$-means clustering, but where
$K$ can vary with the number of observations $N$. This \textit{DP-means}
algorithm mirrors the well-known \emph{small variance asymptotic}
derivation of the $K$-means algorithm from a simplification of the
\emph{expectation-maximization }(E-M)\emph{ }algorithm for the finite
Gaussian mixture model \citep[page 423]{Bishop2006prml}.

In DP-means, each Gaussian component is spherical with identical covariance
$\Sigma_{k}=\sigma\mathbf{I}$, and the variance parameter $\sigma>0$
is assumed known and hence fixed in the algorithm. This is equivalent
to assuming $\R_{k}=\sigma^{-1}\mathbf{I}$ in the DPGMM model (\ref{eq:generative model DPGMM-1}).
Then, since the cluster components have fixed covariances, the conjugate
choice for the cluster means is Gaussian. To obtain closed form solutions
\citet{ICML2012Kulis} assume a zero mean Gaussian prior with covariance
$\rho\mathbf{I}$ and fixed $\rho>0$. They further assume a functional
dependency between $\alpha$ and the covariances, $\alpha=\sqrt{1+\frac{\rho}{\sigma}}\cdot\exp\left(-\frac{\lambda}{2\sigma}\right)$,
for some new parameter $\lambda>0$. The probability of assigning
observation $i$ to cluster $k$ becomes:

\begin{equation}
p\left(z_{i}^{j}=k\left|\boldsymbol{\mu}_{k}^{j-1},\mathbf{\Sigma}_{k}^{j-1}\right.\right)\propto N_{k,-i}^{j-1}\exp\left(-\frac{1}{2\sigma}\left\Vert \boldsymbol{\x}_{i}-\boldsymbol{\mu}_{k}\right\Vert _{2}^{2}\right)
\end{equation}

and the probability for creating a new cluster is:

\begin{equation}
p\left(z_{i}^{j}=K+1\left|G_{0}\right.\right)\propto\exp\left(-\frac{1}{2\sigma}\left[\lambda+\frac{\sigma}{\rho+\sigma}\left\Vert \boldsymbol{\x}_{i}\right\Vert _{2}^{2}\right]\right)
\end{equation}

Then, in the small variance asymptotic limit $\sigma\rightarrow0$
(as in $K$-means) the probability over $z_{i}=k$ collapses to $1$
when $\boldsymbol{\mu}_{k}$ has the smallest distance to $\boldsymbol{\x}_{i}$;
or instead, the probability of creating a new cluster becomes $1$
when $\lambda$ is smaller than any of these distances. Therefore,
a new cluster is created if there are any observed data points for
which $\lambda$ is smaller than the distance from that data point
to any existing component mean vector. If a new component is generated,
it will have $\boldsymbol{\mu}_{k+1}=\boldsymbol{\x}_{i}$ because
in the small variance limit, the covariance of the posterior over
$\boldsymbol{\mu}_{k+1}$ becomes zero.

The component parameter update stage simplifies to the $K$-means
update, i.e. the means of each component are simply replaced by the
mean of every observation assigned to that component. This occurs
because by conjugacy the posterior over the component means is multivariate
Gaussian and as $\sigma\to0$ the likelihood term dominates over the
prior.

\section{MAP-DPM using elliptical multivariate Gaussians\label{sec:MAP-DP-description}}

Although the DP-means algorithm presented above is straightforward,
it has various drawbacks in practice. The most problematic is that
the functional dependency between the concentration parameter and
the covariances destroys the preferential attachment property of the
DPM because the counts of assignments to components $N_{k,-i}$ no
longer influence which component gets assigned to an observed data
point. Only the geometry in the data space matters. A new cluster
is created by comparing the parameter $\lambda$ against the distances
between cluster centers and data points so that the number of clusters
is controlled by the geometry alone, and not by the number of data
points already assigned to each cluster. So, for high-dimensional
datasets, it is not clear how to choose the parameter $\lambda$.
By contrast, in the CRP Gibbs sampler for the DPM, the concentration
parameter controls the rate at which new clusters are produced in
a way which is largely independent of the geometry.

Another problem with small variance asymptotics is that the introduction
of degenerate Dirac point masses causes likelihood comparisons to
be no longer meaningful since the model likelihood becomes infinite.
This means that we cannot readily choose parameters such as $\lambda$
using standard model selection methods such as cross-validation.

Here, we propose a DPM inference algorithm based on \emph{iterated
conditional modes} (ICM, see \citet{kittler1984contextual} and also
\citet[page 546]{Bishop2006prml}). This is also called the \emph{maximization-maximization
}(M-M)\emph{ }algorithm by \citet{Welling07bayesiank-means}. The
basic idea is to use conditional modal point estimates rather than
samples from the conditional probabilities used in Gibbs.

\subsection{Probabilistic model overview\label{sub:Model-overview}}

We will make the simplifying assumption that $\mathbf{\R}_{k}$ are
diagonal, denoting the non-zero entries $\tau_{k,d}$. So, the component
distributions are a product of univariate Gaussians $\mathcal{N}\left(\boldsymbol{\x}_{i}\left|\mu_{k},\mathbf{\R}_{k}^{-1}\right.\right)=\prod_{d=1}^{D}\mathcal{N}\left(x_{i,d}\left|\mu_{k,d},\tau_{k,d}^{-1}\right.\right)$.
Then we can replace the NW prior over the cluster parameters with
the simpler product of \emph{normal-Gamma }(NG) priors, retaining
conjugacy (Appendix \ref{sub:Model-and-prior}). The prior for a specific
cluster component is therefore:
\begin{eqnarray}
g\left(\mu_{k,d},\tau_{k,d}\right) & = & \mathcal{N}\left(\mu_{k,d}\left|m_{0,d},\left(c_{0}\tau_{k,d}\right)^{-1}\right.\right)\\
 & \times & \text{Gamma}\left(\tau_{k,d}\left|a_{0},b_{0,d}\right.\right)\nonumber 
\end{eqnarray}

The NG prior parameters $\thp=\left(\m_{0},c_{0},\b_{0},a_{0}\right)$
need to be specified. As we are only interested in clustering, we
will integrate out the cluster parameters (a process known as \emph{Rao-Blackwellization
}\citep{blackwell1947}\textbf{ }which often leads to more efficient
samplers \citep{Cassella1994}). In the Gibbs sampler framework, this
requires integrating out the cluster parameters from the conditional
of the cluster indicators given the cluster parameters, $p\left(z\left|\boldsymbol{\mu}_{k},\R{}_{k}\right.\right)$.
In the DPGMM Gibbs sampler, we assign observation $i$ to cluster
$k$ with probability $p\left(z_{i}=k\left|\boldsymbol{\mu}_{k},\R_{k}\right.\right)\varpropto N_{k,-i}\mathcal{N}$$\left(\boldsymbol{\x}_{i}\left|\boldsymbol{\mu}_{k},\R_{k}\right.\right)$.
The integration $p\left(z_{i}=k\left|z_{-i}\right.\right)\varpropto N_{k,-i}\int\mathcal{N}\left(\boldsymbol{\x}_{i}\left|\boldsymbol{\mu}_{k},\R_{k}\right.\right)p\left(\boldsymbol{\mu},\R\right)d\boldsymbol{\mu}d\R$
is tractable because of conjugacy. Integrating out the means and precisions
from the corresponding Gaussian likelihoods of observation $\x_{i}$,
we obtain the cluster assignment probabilities for assigning a point
to an existing cluster $k\in\{1,\ldots,K\}$ or a new cluster $K+1$,
respectively: 
\begin{eqnarray}
p\left(z_{i}=k\left|z_{-i},\thp\right.\right) & \propto & N_{k,-i}\, p\left(\x_{i}|z_{-i}\right)\label{eq:gibbsexisting}\\
p\left(z_{i}=K+1\left|\conc,\thp\right.\right) & \propto & \alpha\, p\left(\x_{i}|\thp\right)\label{eq:gibbsnew}
\end{eqnarray}

where for brevity we have used the shorthand $p\left(\x_{i}|z_{-i}\right)=p\left(\x_{i}|z_{-i},\x_{-i},\z_{i}=k\right)$.
The right hand side conditional probabilities are: 
\begin{align}
p\left(\x_{i}|z_{-i}\right) & =\prod_{d=1}^{D}\mathrm{\St}\left(x_{i,d}\left|m_{k,d}^{-i},\frac{a_{k}^{-i}c_{k}^{-i}}{b_{k,d}^{-i}\left(c_{k}^{-i}+1\right)},2a_{k}^{-i}\right.\right)\label{eq:stmarginal_ex}\\
p\left(\x_{i}|\thp\right) & =\prod_{d=1}^{D}\mathrm{\St}\left(x_{i,d}\left|m_{0,d},\frac{a_{0}c_{0}}{b_{0,d}\left(c_{0}+1\right)},2a_{0}\right.\right)\label{eq:stmarginal_new}
\end{align}

Here, $\mathrm{\St}\left(x_{i,d}\left|\mu,\Lambda,\nu\right.\right)$
denotes a Student-T distribution with mean $\mu$, precision $\Lambda$
and degrees of freedom $\nu$, and $\left(\m_{k}^{-i},c_{k}^{-i},\b_{k}^{-i},a_{k}^{-i}\right)$
are the NG posterior parameters -- which we call \emph{component statistics}.
For cluster $k$ and dimension $d$, after removing the effect of
the current observation \citep{neal2000markov}:
\begin{align}
m_{k,d}^{-i} & =\frac{c_{0}m_{0,d}+N_{k,-i}\bar{x}_{k,d}^{-i}}{c_{0}+N_{k,-i}}\nonumber \\
c_{k}^{-i} & =c_{0}+N_{k,-i}\nonumber \\
a_{k}^{-i} & =a_{0}+N_{k,-i}/2\label{eq:Sufficient statistics def}\\
b_{k,d}^{-i} & =b_{0,d}+\frac{1}{2}\sum_{j:z_{i}=k,j\neq i}\left(x_{j,d}-\bar{x}_{k,d}^{-i}\right)^{2}\nonumber \\
 & +\frac{c_{0}N_{k,-i}\left(\bar{x}_{k,d}^{-i}-m_{0,d}\right)^{2}}{2\left(c_{0}+N_{k,-i}\right)}\nonumber 
\end{align}

Note that these component statistics can be efficiently computed by
adding and removing the effect of a single observation from the statistics
of each cluster, $\sum_{i:z_{i}=k}x_{i,d}$ and $\sum_{i:z_{i}=k}x_{i,d}^{2}$
(see Appendix \ref{sec:Cluster-assignments}, Algorithm \ref{alg.mapdpFast}).

The degrees of freedom of the Student-T marginal likelihoods $p\left(\x_{i}|z_{-i}\right)$
depend upon the number of observations in the corresponding component.
The likelihood for smaller components will have heavier than Gaussian
tails, while the likelihood for clusters assigned a large number of
observations will be close to Gaussian. This makes the clustering
more robust to outliers and penalizes creating clusters with few observations
assigned to them.

In summary, the nonparametric Bayesian probabilistic model we will
use is (see Figure \ref{fig.graphicalModel}): 
\begin{eqnarray}
\zeta_{N} & \sim & \textrm{CRP}\left(\conc,N\right)\label{eq:Stundent-T generating model}\\
x_{i,d} & \sim & \textrm{\ensuremath{\mathrm{\St}}}\left(\mu_{z_{i},d},\Lambda_{z_{i},d},\nu_{z_{i}}\right)\nonumber 
\end{eqnarray}

\begin{center}

%
%
%
%
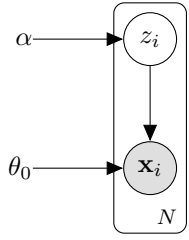
\begin{figure}[h]
\begin{center}
\begin{tikzpicture}
\node[latent] (z) {$\z_i$};
\node[obs, below=of z] (x) {$\x_i$};

\node[const, left=1.2cm of z] (a) {$\conc$};
\node[const, left=1.2cm of x] (theta0) {$\thp$};

\edge {z} {x} ; %
\edge {a} {z} ; %
\edge {theta0} {x} 

\plate {zx} {(z)(x)} {$N$} ;
\end{tikzpicture}
\end{center}
\caption{Collapsed DPM graphical model. Inference in this probabilistic model is performed using the MAP-DPM algorithm described in the text.}
\label{fig.graphicalModel}
\end{figure}

\par\end{center}

\subsection{MAP-DPM inference algorithm\label{sub:MAP-DPM-inference-algorithm}}

MAP inference applied to the probabilistic model above involves finding
the mode for each individual Gibbs step described in (\ref{eq:gibbsexisting})-(\ref{eq:gibbsnew}).
For observation $\x_{i}$, we compute the negative log probability
for each existing cluster $k$ and for a new cluster $K+1$:
\begin{equation}
\begin{alignedat}{2}q_{i,k} & = & -\log\Nkn-\log p\left(\boldsymbol{\x}_{i}|z_{-i}\right)\\
q_{i,K+1} & = & -\log\conc-\log p\left(\boldsymbol{\x}_{i}|\thp\right)
\end{alignedat}
\label{eq:negative_log existing & new components}
\end{equation}

omitting quantities independent of $k$ (detailed expressions in the
Appendix \ref{sec:Cluster-assignments}). For each observation $\boldsymbol{\x}_{i}$
we compute the above $K+1$-dimensional vector $\boldsymbol{q}_{i}$
and select the cluster number of the smallest element from it: 
\[
z_{i}=\argmin_{k\in\left\{ 1,\ldots,K,K+1\right\} }q_{i,k}
\]

The algorithm proceeds to the next observation $\boldsymbol{\x}_{i+1}$
by updating the component statistics $\left(\m_{k}^{-\left(i+1\right)},c_{k}^{-\left(i+1\right)},\b_{k}^{-\left(i+1\right)},a_{k}^{-\left(i+1\right)}\right)$
to reflect the new value of $z_{i}$ and remove the effect of data
point $\x_{i+1}$. To check convergence of the algorithm we compute
the complete data likelihood:

\begin{equation}
\begin{split}p\left(\boldsymbol{\x},\z|\conc\right)= & \left(\prod_{i=1}^{N}\prod_{k=1}^{K}p\left(\boldsymbol{\x}_{i}|\z_{-i}\right){}^{\delta\left(z_{i},k\right)}\right)p\left(z_{1},\ldots,z_{N}\right)\end{split}
\end{equation}

where $\delta\left(z_{i},k\right)$ is the Kronecker delta function
and $p\left(z_{1},\ldots z_{N}\right)$ is defined in (\ref{eq:joint over z}).
The negative log of this quantity (\emph{negative log likelihood},
NLL) is used to assess convergence as described in Algorithm \ref{alg.mapdp}.
ICM is guaranteed to never increase the NLL at each iteration step
and therefore the MAP-DPM algorithm will converge to a fixed point
\citep{Welling07bayesiank-means}. The susceptibility of the algorithm
to local minima can be alleviated using multiple restarts with random
parameter initializations. The existence of a closed-form, non-degenerate
likelihood (unlike in small variance asymptotic approaches) can be
used to estimate the model parameters, such as the concentration parameter,
and perform model selection (see Appendix \ref{sec:Learning-the-concentration}).
We can also use techniques such as cross-validation by computing an
out-of-sample likelihood, which we now discuss.

\begin{algorithm}            
\caption{MAP-DPM}          
\label{alg.mapdp}            
\begin{algorithmic}          
	\REQUIRE $\x_1,\ldots,\x_N$: data, $\conc$: concentration, $\epsilon$: threshold, $\thp$ prior.
	\RETURN $\z_1,\ldots,\z_N$: indicators, $K$: clusters.
	\STATE {\bf Initialisation}: $K=1,\z_i = 1$, $\forall i \in \{1,\ldots,N\}$
    \WHILE{likelihood change $\Delta\left[-\log p(\x,\z|\conc)\right] < \epsilon$}
		\FORALL{observations $i \in \{1,\ldots,N\}$}
			\FORALL{existing clusters $k \in \{1,\ldots,K\}$}
				 \STATE Compute $\left( \mathbf{m}_{k}^{-i},c_{k}^{-i},\mathbf{b}_{k}^{-i},a_{k}^{-i} \right)$, equation  $\eqref{eq:Sufficient statistics def}$
				\STATE Compute $q_{i,k}=-\log\Nkn-\log p(\x_{i}|\z^{-i})$ 
			\ENDFOR
			\STATE Compute $q_{i,K+1}=-\log\conc-\log p(\x_{i}|\thp)$
			\STATE Compute $\z_i=\argmin_{k\in\left\{ 1,\ldots,K,K+1\right\} }q_{i,k}$
		\ENDFOR
    \ENDWHILE
\end{algorithmic}
\end{algorithm}

\subsection{Out-of-sample prediction\label{sub:Out-of-sample-prediction}}

To compute the out-of-sample likelihood for a new observation $\boldsymbol{\x}_{N+1}$
we can consider two approaches that differ in how the indicator $z_{N+1}$
is estimated. Firstly, the unknown indicator $z_{N+1}$ may be integrated
out resulting in a mixture density: 
\begin{multline}
p\left(\boldsymbol{\x}_{N+1}|\conc,\z_{N}\right)=\\
\sum_{k=1}^{K+1}p\left(z_{N+1}=k|\z_{N},\conc\right)p\left(\x_{N+1}|\z_{N},z_{N+1}=k\right)
\end{multline}

where we have omitted the dependency on the training observations
$\X_{N}$. The assignment probability $p\left(z_{N+1}=k|z_{N},\conc\right)$
is $\frac{N_{k}}{\conc+N}$ for an existing cluster and $\frac{\conc}{\conc+N}$
for a new cluster. The second term $p\left(\x_{N+1}|z_{N},z_{N+1}=k\right)$
corresponds to the Student-T marginal in (\ref{eq:stmarginal_ex})
and (\ref{eq:stmarginal_new}) for an existing and a new cluster,
respectively. Alternatively, we can use a point estimate for $z_{N+1}$
by picking the minimum negative log posterior of the indicator $p\left(z_{N+1}|\boldsymbol{\x}_{N+1},z_{N},\conc\right)$
or equivalently: 
\begin{multline}
\z_{N+1}^{\mathrm{MAP}}=\argmin_{k\in\left\{ 1,\ldots,K,K+1\right\} }\left[\log p\left(\boldsymbol{\x}_{N+1}|k\right)-\log p\left(k|z_{N},\conc\right)\right]
\end{multline}

The first (marginalization) approach is used in \citet{blei2004variational}
and is more robust as it incorporates the probability mass of all
cluster components. However the second (modal) approach can be useful
in cases where only a point prediction is needed such as when computing
the test set normalised mutual information (see Section \ref{sec:Synthetic-Experiment:CRP}).

\section{Experiments\label{sec:Experiments-on-Synthetic}}

\subsection{Synthetic CRP parameter estimation\label{sec:Synthetic-Experiment:CRP}}

\begin{figure}[htbp]  
\centering
\fbox{
\includegraphics[width=0.5\columnwidth]{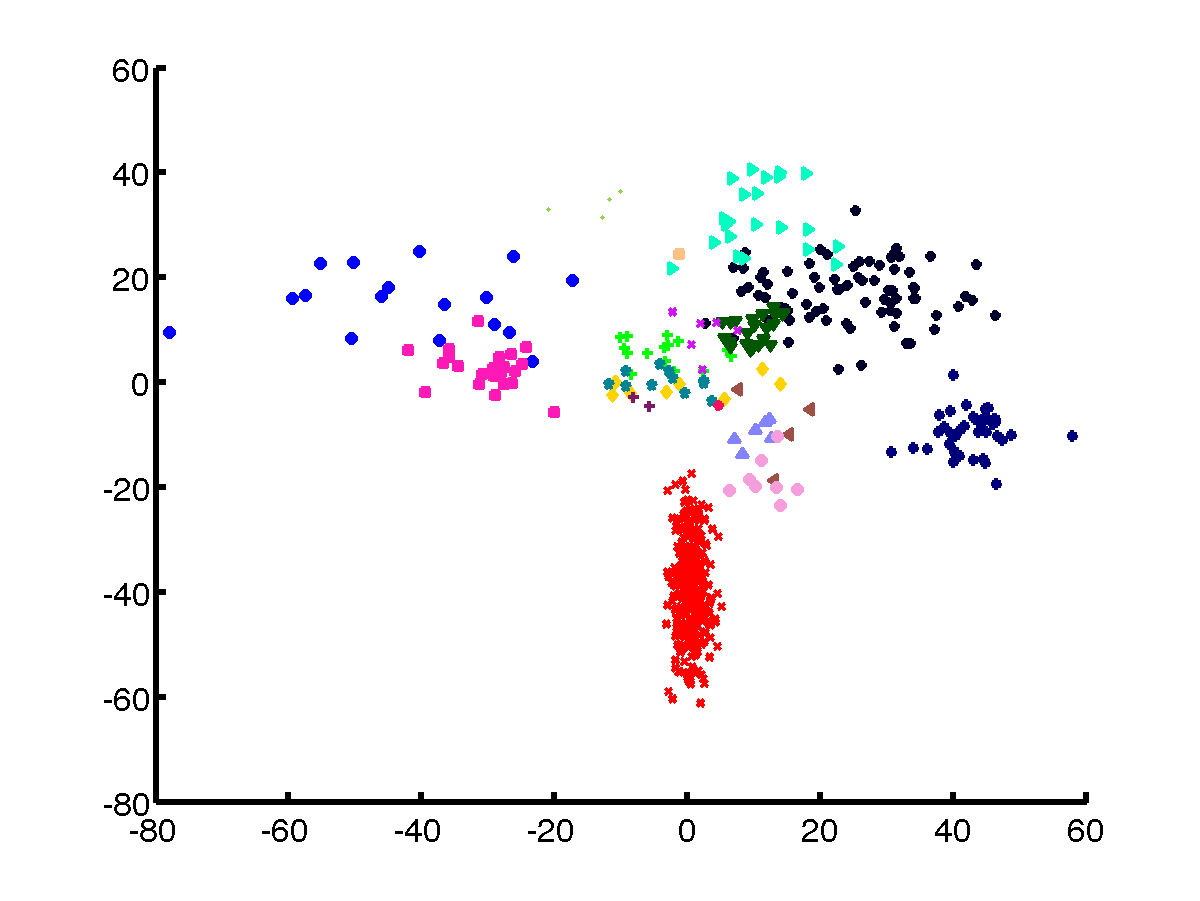}
}
\caption{Sample from $D=2$ CRP probabilistic model containing $K=18$ clusters ranging in size from the biggest cluster $N_k=18$ to two clusters with $N_k=1$.}
\label{fig.ptnsample}
\end{figure}

We examine the performance of the MAP-DPM algorithm in terms of discovering
the partitioning distribution, and the computational effort needed
to do this. We generate 100 samples from a two-dimensional CRP model
(\ref{eq:generative model DPGMM-1}) with diagonal precision matrices
(example in Figure \ref{fig.ptnsample}). The partitioning is sampled
from a CRP with fixed concentration parameter $\conc=3$ and data
size $N=600$. Gaussian component parameters are sampled from an NG
prior with parameters $\m_{0}=[1,1],c_{0}=0.1,\b_{0}=\left[10,10\right],a_{0}=1$.
As expected, when using a CRP prior, the sizes of the different clusters
vary significantly with many small clusters containing only a few
observations in them.

We fit a CRP mixture model with integrated-out component parameters
(\ref{eq:Stundent-T generating model}) by MAP and integrated-out
Gibbs inference, using the known, ground truth model values for the
NG prior and $\alpha$ used to generate the data. Convergence for
the Gibbs algorithm is tested using the Raftery diagnostic ($q=0.025,r=0.1,s=0.95$)
\citet{raftery1992many}. We use a high convergence acceptance tolerance
of $r=0.1$ to obtain less conservative estimates for the number of
iterations required. As there is no commonly accepted way to check
the convergence of MCMC algorithms, our comparison is by necessity
somewhat arbitrary but we believe the choices we have made are realistic
and useful conclusions may be drawn from the comparison.

We measure clustering estimation accuracy using the (sum) \emph{normalized
mutual information} $\text{NMI}(U,V)=\frac{2H(U,V)}{H(U)+H(V)}$ between
the ground truth clustering $U$ and the estimate $V$ \citep{vinh2010information},
where $H$ is the entropy. NMI lies in the range $[0,1]$ with higher
values signifying closer agreement between the clusterings (see Appendix
\ref{sec:CRP-Experiment} for other reported measures, e.g. the \emph{adjusted
mutual information }\citep{vinh2010information}).

\begin{table}
\caption{Performance of MAP-DPM and Gibbs algorithms on the CRP mixture experiment
(Section \ref{sec:Synthetic-Experiment:CRP}). Mean and two standard
deviations (in brackets) reported across the 100 CRP mixture samples.
The quantity$\Delta K$ is the difference between the estimated number
of clusters and the known number of clusters. The range of the normalized
mutual information (NMI) is $[0,1]$ with higher values reflecting
lower clustering error. \label{tab:crptrain}\smallskip{}
}

\centering{}%
\begin{tabular}{|c|c|c|}
\hline 
 & Gibbs  & MAP-DPM\tabularnewline
\hline 
\hline 
NMI & 0.67 (0.23) & 0.71 (0.23)\tabularnewline
\hline 
Iterations & 2,020 (900) & 13.3 (13.9)\tabularnewline
\hline 
CPU time (secs) & 7,300 (4,000) & 28 (32)\tabularnewline
\hline 
$\Delta K$ & -0.91 (7.12) & -6.94 (6.93)\tabularnewline
\hline 
Empty clusters & 5,900 (3,600) & 10.33 (22.1)\tabularnewline
\hline 
Test set NMI & 0.72 (0.20) & 0.71 (0.23)\tabularnewline
\hline 
\end{tabular}
\end{table}

In Table \ref{tab:crptrain} a range of performance metrics for the
MAP-DPM and Gibbs algorithms are shown. Both MAP-DPM and Gibbs achieve
similar clustering performance for both training and test set NMI.
To assess out-of-sample performance, another set of $N=600$ observations
were sampled from each CRP mixture sample and the out-of-sample point
prediction calculated for each model (Section \ref{sub:Out-of-sample-prediction}).
The Gibbs sampler requires, on average, approximately 152 times the
number of iterations to converge than MAP-DPM, as reflected in the
CPU times. Also, the number of empty clusters created in Gibbs is
higher than MAP-DPM due to the higher number of iterations required;
an effective Gibbs implementation therefore would need to efficiently
handle the empty clusters.

\begin{figure}[htbp]  
\centering
\includegraphics[width=0.8\columnwidth]{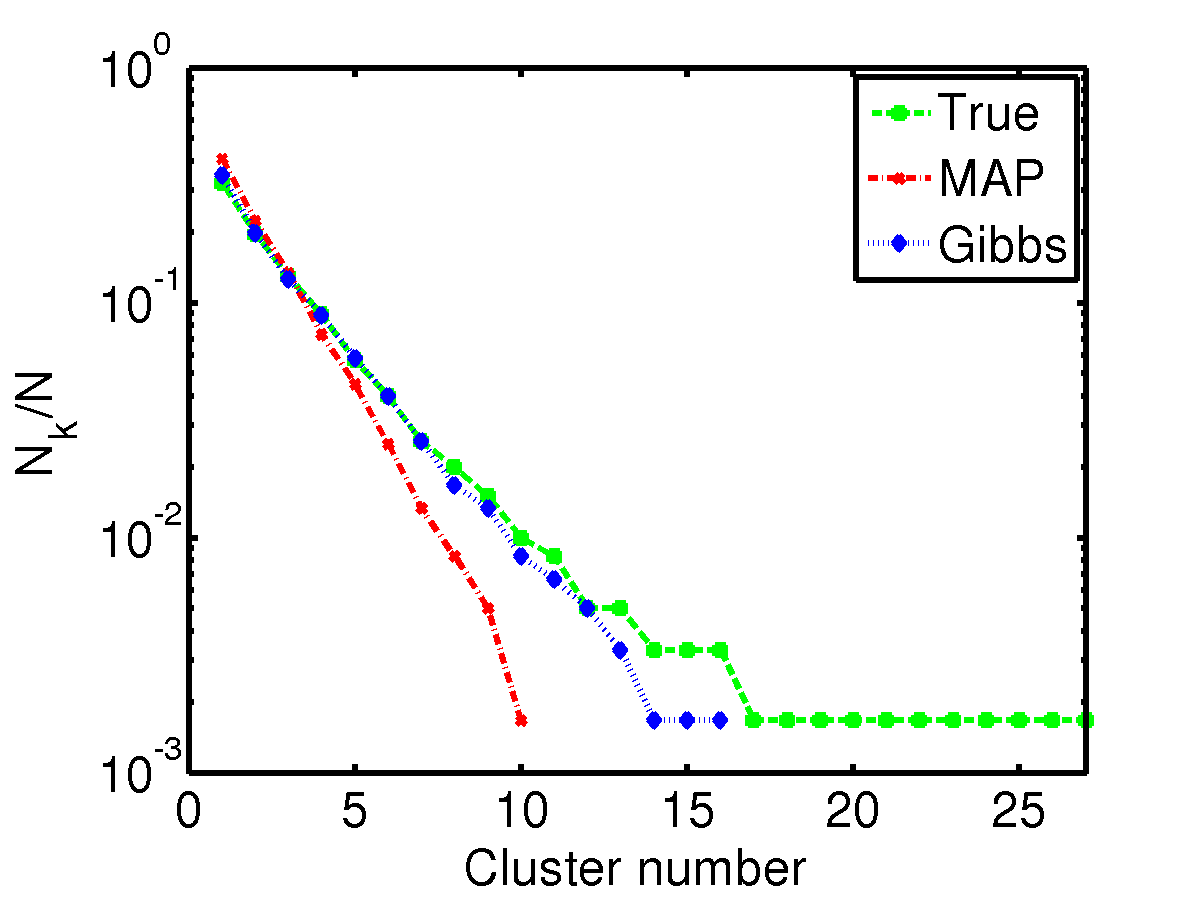}
\caption{Synthetically-generated CRP mixture data experiments; distribution of cluster sizes, actual and estimated using MAP-DPM and  Gibbs. Cluster number ordered by decreasing size (horizontal axis) vs $\frac{N_k}{N}$ (vertical axis, log scale).} 
\label{fig.ptndistr}
\end{figure}

When examining the number of clusters ($\Delta K$), Gibbs is closest
to the ground truth whilst MAP-DPM produces signficant underestimates.
In Figure \ref{fig.ptndistr} the median partitioning is shown in
terms of the partitioning $N_{k}/N$ and the number of clusters. MAP-DPM
fails to identify the smaller clusters whereas the Gibbs sampler is
able to do so to a much greater extent. This is a form of underfitting
where the MAP algorithm captures the mode of the partitioning distribution
but fails to put enough mass on the tails (the smaller clusters);
that may also be described as an underestimation of the variance of
the partitioning distribution. The NMI scores do not reflect this
effect as the impact of the smaller clusters on the overall measure
is minimal.

\subsection{UCI datasets \label{sec:Synthetic-Experiment:UCI}}

We compare the DP-means, MAP-DPM and Gibbs samplers on seven UCI datasets
and assess their performance using the same NMI measure as in Section
\ref{sec:Synthetic-Experiment:CRP}. Class labels in the datasets
are treated as cluster numbers.%
\footnote{We do not assess ``Car'' and ``Balance scale'' datasets used in
\citet{ICML2012Kulis} because they consist of a complete enumeration
of 6 and 4 categorical factors respectively, and it is not meaningful
to apply an unsupervised clustering algorithm to such a setting.%
} As in Section \ref{sec:Synthetic-Experiment:CRP} the Gibbs sampler
is stopped using the Raftery diagnostic \citep{raftery1992many}.
For DP-means, we choose $\lambda$ to give the true number of clusters
in the corresponding dataset \citep{ICML2012Kulis}. The NG prior
in the Gibbs and MAP-DPM algorithms is set empirically; the mean is
set to the sample mean of the data, $c_{0}=\frac{10}{N}$ , a scaled
version of value proposed in \citet{kass1995reference}, $a_{0}=1$
and each dimension of $\b_{0}$ is set to the sample variance of the
data in that corresponding dimension.%
{} Concentration parameter $\alpha$ is selected by minimizing the NLL
(Section \ref{sec:MAP-DP-description}) in the MAP-DPM algorithm across
a discrete set of candidate values and reused in the Gibbs algorithm.
For the Gibbs algorithm, we compute the mean and two standard deviations
for the NMI score across all samples (Table \ref{tab:Clustering-NMI}).

\begin{table}
\caption{Clustering performance of DP-means, MAP-DPM, and Gibbs samplers on
UCI datasets, measured using NMI (two standard deviations in brackets),
averaged over all runs.\label{tab:Clustering-NMI}\smallskip{}
}

\centering{}%
\begin{tabular}{|c|c|c|c|}
\hline 
 & DP-means & Gibbs  & MAP-DPM\tabularnewline
\hline 
\hline 
Wine & 0.42 & 0.72 (0.06) & 0.86\tabularnewline
\hline 
Iris & 0.76 & 0.75 (0.06) & 0.76\tabularnewline
\hline 
Breast cancer & 0.75 & 0.70 (0.01) & 0.71\tabularnewline
\hline 
Soybean & 0.36 & 0.49 (0.00) & 0.40\tabularnewline
\hline 
Parkinson's & 0.02 & 0.13 (0.02) & 0.12\tabularnewline
\hline 
Pima & 0.03 & 0.07 (0.01) & 0.07\tabularnewline
\hline 
Vehicle & 0.21 & 0.15 (0.02) & 0.15\tabularnewline
\hline 
\end{tabular}
\end{table}

On almost all of the datasets (6 out of 7), MAP-DPM is comparable
to, or even better than, the Gibbs sampler, and on 5 out 7 datasets
it performs better than DP-means (Table \ref{tab:Clustering-NMI}).
DP-means performs well on lower-dimensional datasets with a small
number of clusters. In higher dimensions, it is more likely for the
clusters to be elliptical rather than spherical and in such cases
the other algorithms outperform DP-means because of the more flexible
model assumptions. In addition, for higher dimensional data it is
more often the case that the different features have different numerical
scales, so the squared Euclidean distance used in DP-means is inappropriate.
Furthermore, MAP-DPM and the Gibbs sampler are more robust to smaller
clusters due to the longer tails of the Student-T distribution and
the rich-get-richer effect of existing clusters assigned many observations.
DP-means is especially sensitive to geometric outliers and can easily
produce excessive numbers of spurious clusters for poor choices of
$\lambda$.

Even though MAP-DPM only gives a point estimate of the full Gibbs
distribution, MAP-DPM can in practice achieve higher NMI scores. This
can occur both because of Gibbs convergence issues, and because we
take the average NMI across all Gibbs samples, where MAP-DPM could
correspond to a Gibbs sample with much higher NMI than average. For
instance, in the Wine dataset (178 observations, 13 dimensions), the
NMI across different Gibbs samples varies considerably and the MAP-DPM
NMI score is close to the highest one achieved across all Gibbs samples.
In the Soybean dataset (266 observations, 35 dimensions), visual inspection
of the Gibbs samples revealed slow Markov chain mixing and even after
10,000 iterations, the samples had not converged. The sparseness of
the data in such a high-dimensional space makes this a particularly
challenging clustering problem and a more sophisticated MCMC sampling
method would likely be required in practice.

We emphasize that these algorithms attempt to maximize the model fit
rather than maximize NMI. The true labels would not be available in
practice and it is not always the case that maximizing the likelihood
also maximizes NMI. Furthermore, if we choose the NG parameters for
each dataset separately, by minimizing the negative log likelihood
with respect to each parameter, higher NMI can been achieved, but
choosing empirical estimates for the model parameters simplifies the
computations.

\begin{table}
\caption{Iterations required to achieve convergence for the DP-means and MAP-DPM
algorithm, and the Gibbs sampler, on datasets from the UCI repository.
`+' indicates convergence was not obtained.\label{tab:Iterations}\smallskip{}
}

\centering{}%
\begin{tabular}{|c|c|c|c|}
\hline 
 & DP-means & Gibbs  & MAP-DPM\tabularnewline
\hline 
\hline 
Wine & 19 & 2,365 & 11\tabularnewline
\hline 
Iris & 8 & 1,543 & 5\tabularnewline
\hline 
Breast cancer & 8 & 939 & 8\tabularnewline
\hline 
Soybean & 14 & 10,000+ & 9\tabularnewline
\hline 
Parkinson's & 1,000+ & 1,307 & 13\tabularnewline
\hline 
Pima & 20 & 1,189 & 17\tabularnewline
\hline 
Vehicle & 12 & 939 & 9\tabularnewline
\hline 
\end{tabular}
\end{table}

In all cases, the MAP-DPM algorithm converges more rapidly than the
other algorithms (Table \ref{tab:Iterations}). The Gibbs sampler
takes, on average, approximately $150$ more iterations to converge
than MAP-DPM to achieve comparable NMI scores. The computational complexity
per iteration for Gibbs and MAP-DPM is comparable, requiring the computation
of the same quantities. This makes the Gibbs sampler significantly
less efficient than MAP-DPM in finding a good labeling for the data.
The price per iteration for DP-means can often be considerably smaller
than MAP-DPM or the Gibbs sampler, as one iteration often does not
include a sweep through all of the data points. This occurs because
the sweep ends when a new cluster has to be created, unlike MAP-DPM
and Gibbs. But, this also implies that DP-means requires more iterations
to converge than MAP-DPM.

\section{Discussion and future directions\label{sec:Discussion-and-Extensions}}

We have presented a simple algorithm for inference in DPMs based on
non-degenerate MAP, and demonstrated its efficiency and accuracy by
comparison to the ubiquitous Gibbs sampler, and a simple alternative,
the small variance asymptotic approach. We believe our approach is
highly relevant to applications since, unlike the small variance approach,
it retains the preferential attachment (rich-get-richer) property
while needing two orders of magnitude fewer iterations than Gibbs.
Unlike the asymptotic approach, an out-of-sample likelihood may be
computed allowing the use of standard model selection and model fit
diagnostic procedures. Lastly, the non-degenerate MAP approach requires
no factorization assumptions for the model distribution unlike VB.

As with all MAP methods, the algorithm can get trapped in local minima,
however, standard heuristics such as multiple random restarts can
be employed to mitigate the risk. This would increase the total computational
cost of the algorithm somewhat but even with random restarts it would
still be far more efficient than the Gibbs sampler.

Although not reported here due to space limitations, we point out
that different implementations of the Gibbs sampler can lead to different
MAP inference algorithms for different model DPMs which naturally
arise from this probabilistic graphical model structure \citep{neal2000markov}.
In general, we have found the resulting alternative algorithms to
be less robust in practice, as they retain the Gaussian likelihood
over the observations given the cluster indicators. If such assumptions
are justified, however, then our MAP approach can be readily applied
to these models as well, for example, where non-conjugate priors are
appropriate.

The generality and the simplicity of our approach makes it reasonable
to adapt to other Bayesian nonparametric mixture models, for example
the Pitman-Yor process which generalizes the CRP \citep{Pitman97thetwo-parameter}.
The MAP approach can also be readily applied to hierarchical Bayesian
nonparametric models such as the Hierarchical DP \citep{teh2006hierarchical}
and nested DP \citep{rodriguez2008nested}. Another useful direction,
for large-scale datasets in particular, would be to extend our approach
to perform inference that does not need to sweep through the entire
dataset in each iteration, for increased efficiency \citep{welling2011bayesian}.

\bibliographystyle{unsrtnat}
\phantomsection\addcontentsline{toc}{section}{\refname}\bibliography{bibFile}

\appendix

\section{Appendix}

\subsection{Model and prior\label{sub:Model-and-prior}}

The mixture model given the component means, precisions and weights
for each component, is:

\begin{equation}
p\left(\x_{i}\right)=\sum_{k=1}^{K}\pi_{k}\N_{D}\left(\x_{i}|\boldsymbol{\vmu}_{k},\R_{k}^{-1}\right)
\end{equation}

where $K$, the number of clusters, becomes infinity in the DPM model.
Variable $\boldsymbol{\vmu}_{k}$ is a $D$-dimensional vector of
cluster means for cluster $k$, and $\R_{k}$ is a diagonal matrix
of precisions, $\tau_{k,d}$. This makes the covariance diagonal,
so the normal can be written as a product of univariate normals, $\N_{D}\left(\x_{i}|\boldsymbol{\vmu}_{k},\R_{k}^{-1}\right)$=
$\prod_{d=1}^{D}\N_{1}\left(x_{i,d}|\vmu_{k,d},(\tau_{k,d})^{-1}\right)$.

We place the conjugate normal-Gamma prior over each dimension of the
cluster means and variances:

\begin{equation}
\begin{alignedat}{2}g(\vmu_{k,d},\tau_{k,d}) & = & \N_{1}\left(\vmu_{k,d}|m_{0,d},(c_{0}\tau_{k,d})^{-1})\right)\text{Gamma}\left(\tau_{k,d}|a_{0},b_{0,d}\right)\\
 & = & \text{NG}\left(\m_{0},c_{0},\b_{0},a_{0}\right)
\end{alignedat}
\end{equation}

Then, we integrate out the cluster parameters of the model, which
is analytically tractable because of the conjugate priors. This results
in a collapsed model structure that places a mixture of Student-T
likelihoods for the probability over an observation:
\begin{equation}
p\left(\x_{i}\right)=\sum_{k=1}^{K}\pi_{k}\prod_{d=1}^{D}\text{St}\left(x_{i,d}|\mu_{k,d},\Lambda_{k,d},df_{k}\right)
\end{equation}

where the parameters of the Student-T are functions of the normal-Gamma
posteriors defined in the paper $\left(\m_{k}^{-i},c_{k}^{-i},\b_{k}^{-i},a_{k}^{-i}\right)$
with effect of observation $\x_{i}$ removed from them. Recall that
$\mu_{k,d}=m_{k,d}^{-i}$, $\Lambda_{k,d}=\frac{a_{k}^{-i}c_{k}^{-i}}{b_{k,d}^{-i}\left(c_{k}^{-i}+1\right)}$
and $df_{k}=2a_{k}^{-i}$. The removal of the effect of the current
observation in the corresponding conditionals occurs because a dependency
is introduced between the cluster indicators, which is a result of
integrating out the cluster parameters (see Algorithm 3 in \citet{neal2000markov}).

\subsection{Cluster assignments \label{sec:Cluster-assignments}}

To complete the MAP algorithm we take the negative logarithm of the
assignment probabilities to arrive at (\ref{eq:negative_log existing & new components}).
For an observation $\x_{i}$, we compute the negative log probability
for each existing cluster, ignoring constant terms: 
\begin{multline}
q_{i,k}=D\log\frac{\Gamma\left(a_{k}^{-i}\right)}{\Gamma\left(a_{k}^{-i}+\frac{1}{2}\right)}-\frac{D}{2}\log\left(\frac{c_{k}^{-i}}{c_{k}^{-i}+1}\right)+\frac{1}{2}\sum_{d=1}^{D}\log b_{k,d}^{-i}\\
+\left(a_{k}^{-i}+\frac{1}{2}\right)\sum_{d=1}^{D}\log\left(1+\frac{c_{k}^{-i}}{2b_{k,d}^{-i}(c_{k}^{-i}+1)}\left(x_{i,d}-m_{k,d}^{-i}\right)^{2}\right)\\
-\log N_{k,-i}\label{eq:negative_log existing component-1}
\end{multline}

Similarly, for a new cluster:
\begin{multline}
q_{i,K+1}=D\log\frac{\Gamma\left(a_{0}\right)}{\Gamma\left(a_{0}+\frac{1}{2}\right)}-\frac{D}{2}\log\left(\frac{c_{0}}{c_{0}+1}\right)+\frac{1}{2}\sum_{d=1}^{D}\log b_{0,d}^{-i}\\
+\left(a_{0}+\frac{1}{2}\right)\sum_{d=1}^{D}\log\left(1+\frac{c_{0}}{2b_{0,d}(c_{0}+1)}\left(x_{i,d}-\mu_{0,d}\right)^{2}\right)\\
-\log\alpha
\end{multline}

For each obsevation $\x_{i}$ we compute the above $K+1$ dimensional
vector $\boldsymbol{q}_{i}$ and select the index of the smallest
element from it:
\begin{equation}
z_{i}=\argmin_{k\in\left\{ 1,\ldots,K,K+1\right\} }q_{i,k}
\end{equation}

In Algorithm \ref{alg.mapdpFast} a fast update version of the method
is described where the NG statistics $\th_{k}^{-i}=\left(\m_{k}^{-i},c_{k}^{-i},\b_{k}^{-i},a_{k}^{-i}\right)$
are updated by removing the effect of one point rather than processing
the entire data set.

\begin{algorithm}            
\caption{MAP-DPM using fast updates}          
\label{alg.mapdpFast}            
\begin{algorithmic}          
	\REQUIRE $\x_1,\ldots,\x_N$: data, $\conc$: concentration, $\epsilon$: threshold, $\th_0$ prior.
	\RETURN $\z_1,\ldots,\z_N$: indicators, $K$: clusters.
	\STATE {\bf Initialisation}: $K=1$, $\z_i = 1$ for all $i \in \{1,\ldots,N\} $.
    \STATE Sufficient statistics, for the global cluster $S_{1,d} = \sum_{i=1}^N x_{i,d}$, $V_{1,d} = \sum_{i=1}^N x_{i,d}^2$, $N_1=N$

	\WHILE{change in likelihood $\Delta\left[-\log p(\x,\z|\conc,\th)\right] < \epsilon$}
		\FORALL{observations $i \in \{1,\ldots,N\}$}
			\FORALL{existing clusters $k \in \{1,\ldots,K\}$}
				\IF{$\z_i = k$}
					\STATE $S_{-i,d} = S_{k,d} - x_{i,d}$
					\STATE $V_{-i,d} = V_{k,d} - x^2_{i,d}$
					\STATE $N_{-i,k} = N_k - 1$
				\ELSE
					\STATE $S_{-i,d} = S_{k,d}$, $V_{-i,d} = V_{k,d}$, $N_{-i,k} = N_k$
				\ENDIF
				\STATE $a_k^{-i} = a_{0}+N_{k,-i}/2$.
				\STATE $c_{k}^{-i}=c_{0}+N_{k,-i}$
				\STATE $m_{k,d}^{-i} = \frac{c_{0}m_{0,d}+S_{-i,d}}{c_{0}+N_{k,-i}}$
				\STATE $b_{k,d}^{-i} = b_{0,d} + \frac{1}{2} \left[ V_{-i} - \frac{ S_{-i,d}^2}{N_{k,-i}} \right] + \frac{c_{0}N_{k,-i}\left(\frac{1}{N_{k,-i}}S_{-i,d}-m_{0,d}\right)^{2}}{2\left(c_{0}+N_{k,-i}\right)}$

				\STATE Compute $q_{i,k}=-\log\Nkn-\log p(\x_{n}|m_{k,d}^{-i},c_{k}^{-i},b_{k,d}^{-i},a_{k}^{-i})$ 
			\ENDFOR
			\STATE Compute $q_{i,K+1}=-\log\conc-\log p(\x_{i}|\th_{0})$
			\STATE Compute $\z_i=\argmin_{k\in\left\{ 1,\ldots,K,K+1\right\} }q_{i,k}$
			\STATE For observation $i$, update $S_{k,d}$, $V_{k,d}$ and $N_k$ if they are affected by change of $z_i$. These are the sufficient statistics for its previous cluster and for its new one, if $z_i$ has changed.
		\ENDFOR
    \ENDWHILE    
\end{algorithmic}
\end{algorithm}

\subsection{Learning the concentration parameter\label{sec:Learning-the-concentration}}

We have considered the following approaches to infer the concentration
parameter $\alpha$:
\begin{enumerate}
\item Cross-validation. By considering a finite set of values for $\alpha$,
choose the value corresponding to the minimum, average out-of-sample
likelihood across all cross-validation folds. This approach is taken
in \citet{blei2004variational} to compare different inference methods.
\item Multiple restarts. Compute the NLL at convergence for each different
value of $\alpha$, and pick the $\alpha$ corresponding to the smallest
NLL at convergence. This is the approach taken in the UCI experiment
section of the paper.
\item MAP estimate. Compute the posterior over $\alpha$ and numerically
locate the mode. An example posterior is given in \citet{rasmussen1999infinite}:
\begin{equation}
p\left(\conc|N,K\right)\propto\frac{\Gamma(\conc)}{\Gamma(\conc+N)}\conc^{K-\frac{3}{2}}\exp\left[-\frac{1}{2\conc}\right]
\end{equation}
where $K$ is the number of existing clusters. This posterior arises
from the choice of an inverse Gamma prior on the concentration parameter
$p(\conc)=IG(1/2,1/2)$ and the likelihood $p\left(N_{1},\ldots,N_{K}|\conc\right)=\frac{\conc^{K}\Gamma(\conc)}{\Gamma(\conc+N)}$.
We generalize this calculation by using a Gamma prior $p(\conc)=\text{Gamma}(a_{\conc},b_{\conc})$:
\begin{equation}
p\left(\conc|N,K\right)\propto\frac{\Gamma(\conc)}{\Gamma(\conc+N)}\conc^{K+a_{\conc}-1}\exp\left[-b_{\conc}\conc\right]
\end{equation}
We numerically minimize the negative log of this posterior using Newton's
method. To ensure the solution is positive we compute the gradient
with respect to $\log\conc$: as \citet{rasmussen1999infinite} notes
$p\left(\log\conc|N,K_{\text{eff}}\right)$ is log-concave and therefore
has a unique maximum (so that the negative log of this has a unique
minimum), where $K_{\text{eff}}$ is the number of non-zero represented
components.
\end{enumerate}
In practice we found the third approach least effective due to the
presence of local minima when doing MAP estimation. The second approach
is the simplest to apply in practice but can be prone to overfitting
for small datasets where we recommend using the cross-validation approach.

\subsection{CRP experiment\label{sec:CRP-Experiment}}

We provide more details on the CRP experiment presented in the paper.
In Tables \ref{tab:crptrain-1}-\ref{tab:crptest} we include the
maximum NMI and \textit{adjusted mutual information} (AMI) score which
corrects for chance effects when comparing clusterings by penalizing
partitions with larger numbers of clusters \citep{vinh2010information}.
To assess out-of-sample accuracy we also include the average, leave-one-out
negative log likelihood discussed in Section \ref{sub:Out-of-sample-prediction}
of the paper in Table \ref{tab:crptest}. All metrics are similar
for the Gibbs and MAP-DPM algorithms, reflecting the good clustering
performance of the MAP algorithm.

\begin{table}
\caption{Performance of MAP-DPM and Gibbs algorithms on the CRP mixture experiment.
Mean and two standard deviations (in brackets) reported across the
100 CRP-mixture samples.\label{tab:crptrain-1} \smallskip{}
}

\centering{}%
\begin{tabular}{|c|c|c|}
\hline 
Measure & Gibbs  & MAP\tabularnewline
\hline 
\hline 
NMI sum & 0.67 (0.23) & 0.71 (0.23)\tabularnewline
\hline 
NMI max & 0.64 (0.24) & 0.64 (0.27)\tabularnewline
\hline 
AMI & 0.62 (0.25) & 0.62 (0.27)\tabularnewline
\hline 
Iterations & 2,000 (900) & 13 (14)\tabularnewline
\hline 
CPU Time (secs) & 7,300 (4,000) & 28 (32)\tabularnewline
\hline 
$\Delta K$ & -0.91 (7.1) & -6.9 (6.9)\tabularnewline
\hline 
Empty clusters & 6,000 (3,600) & 14 (28)\tabularnewline
\hline 
\end{tabular}
\end{table}

\begin{table}
\caption{Test set performance for MAP-DPM and Gibbs algorithms on samples until
convergence. Mean and two standard deviations (in brackets) reported.\label{tab:crptest}\smallskip{}
}

\centering{}%
\begin{tabular}{|c|c|c|}
\hline 
Measure & Gibbs  & MAP\tabularnewline
\hline 
\hline 
NMI sum & 0.72 (0.20) & 0.71 (0.23)\tabularnewline
\hline 
NMI max & 0.68 (0.22) & 0.64 (0.27)\tabularnewline
\hline 
AMI & 0.67 (0.23) & 0.62 (0.27)\tabularnewline
\hline 
NLL (leave-one-out) & 7.17 (1.10) & 7.20 (1.11)\tabularnewline
\hline 
\end{tabular}
\end{table}

\begin{figure}[htbp]  
\centering
\mbox {
\subfigure[5th quantile]{\includegraphics[width=.5\columnwidth]{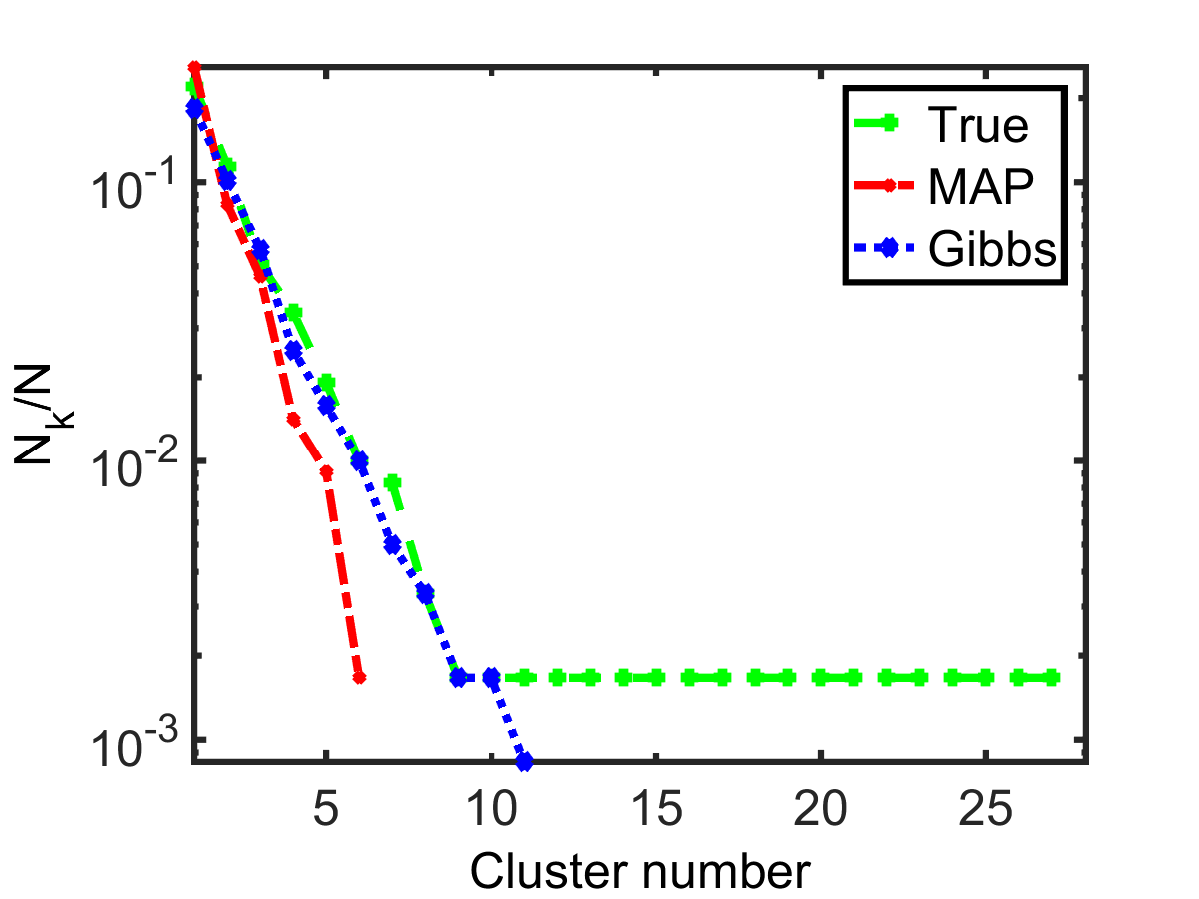}}
\subfigure[95th quantile]{\includegraphics[width=.5\columnwidth]{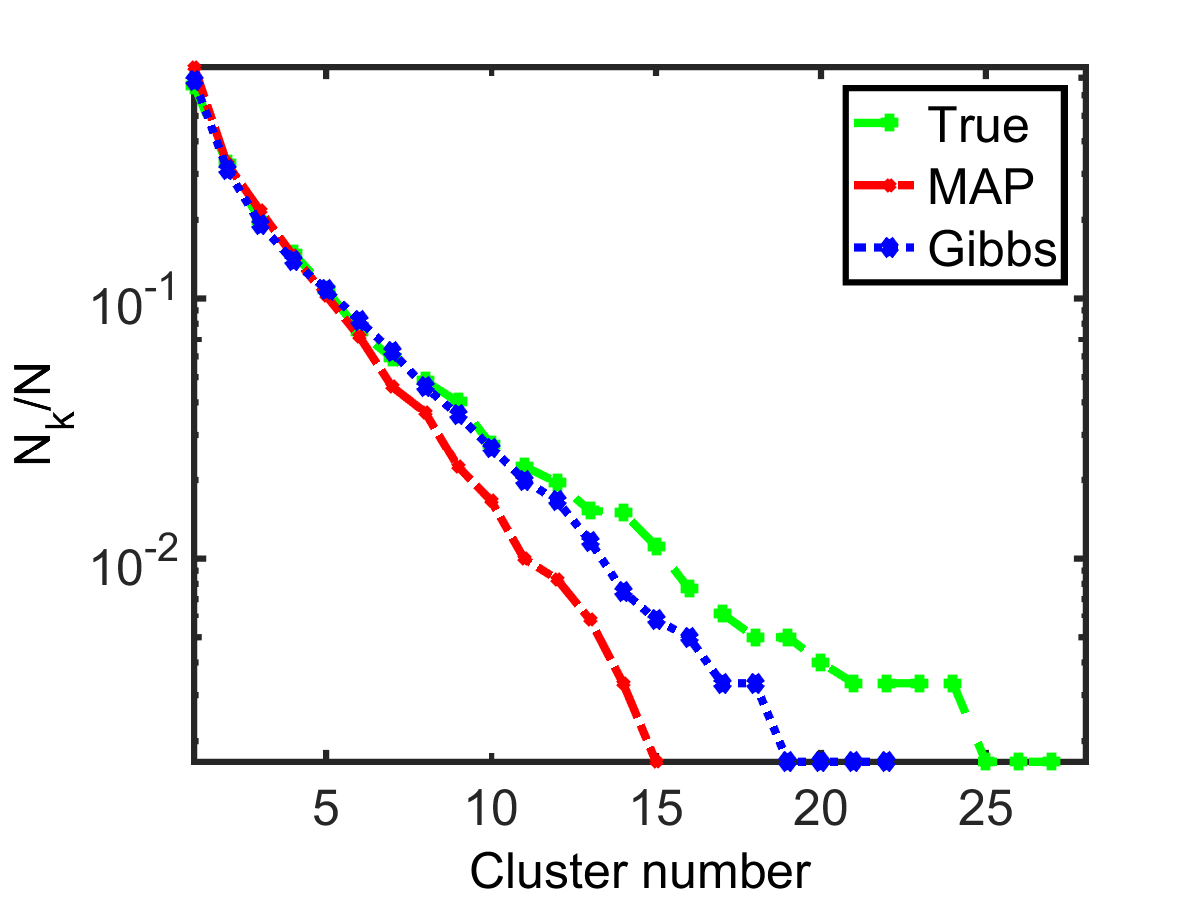}}
}
\caption{Partitioning distribution: 5th and 95th quantiles. Cluster number (horizonal axis) ordered by descreasing cluster size vs $\frac{N_k}{N}$ (vertical axis, log scale).} \label{fig.ptndistrQuantiles}
\end{figure}

In Figure \ref{fig.ptndistrQuantiles} we also show the 5th and 95th
quantiles of the clustering distribution for the CRP ground-truth,
MAP and Gibbs algorithms. We see that the effect discussed in the
paper is present at both extreme quantiles with the MAP algorithm
consistently underestimating the total number of clusters by not identifying
the smaller clusters.

\end{document}